\pdfoutput=1
\documentclass[runningheads]{llncs}
\usepackage{graphicx}
\usepackage{xspace}
\usepackage{amsmath,amssymb} 
\usepackage[pagebackref=true,breaklinks=true,letterpaper=true,colorlinks,bookmarks=false]{hyperref}

\usepackage{color}
\usepackage[width=122mm,left=12mm,paperwidth=146mm,height=193mm,top=12mm,paperheight=217mm]{geometry}

\usepackage{times}
\usepackage{epsfig}
\usepackage{enumitem}
\usepackage{algorithm}
\usepackage{multirow}
\usepackage{subfig}
\usepackage[noend]{algorithmic}
\usepackage{epstopdf}
\usepackage{amsfonts}

\newcommand{\argmin}[1]{\underset{#1}{\operatorname{argmin}}}

\newcommand{\bA}{\mathbf{A}}
\newcommand{\bB}{\mathbf{B}}
\newcommand{\bC}{\mathbf{C}}

\newcommand{\bE}{\mathbf{E}}

\newcommand{\bH}{\mathbf{H}}
\newcommand{\bI}{\mathbf{I}}

\newcommand{\bK}{\mathbf{K}}

\newcommand{\bW}{\mathbf{W}}
\newcommand{\bX}{\mathbf{X}}
\newcommand{\bY}{\mathbf{Y}}

\newcommand{\bc}{\mathbf{c}}

\newcommand{\bk}{\mathbf{k}}

\newcommand{\real}{\mathbb{R}}

\long\def\ignorethis#1{} 
\usepackage{amsfonts}
\usepackage{amsmath}

\newsavebox{\savepar}

\newcommand{\trans}[1]{{#1}^{\ensuremath{\mathsf{T}}}}

\newcommand{\sign}[0]{\operatorname{sign}\xspace}
\newcommand{\argmax}[1]{\underset{#1}{\operatorname{argmax}}}

\def\eg{{\em e.g.,}}

\begin{document}

\pagestyle{headings}
\mainmatter
\def\ECCV16SubNumber{1689}  

\title{XNOR-Net: ImageNet Classification Using Binary Convolutional Neural Networks} 


\authorrunning{Rastegari et al.}

\author{Mohammad Rastegari$^\dag$, Vicente Ordonez$^\dag$, Joseph Redmon$^*$, Ali Farhadi$^{\dag *}$}

\institute{Allen Institute for AI$^\dag$, University of Washington$^*$ \\ \texttt{\footnotesize \{mohammadr,vicenteor\}@allenai.org \\ \{pjreddie,ali\}@cs.washington.edu}}

\maketitle

\begin{abstract}
We propose two efficient approximations to standard convolutional neural networks: Binary-Weight-Networks and XNOR-Networks. In Binary-Weight-Networks, the filters are approximated with binary values resulting in $32\times$ memory saving. In XNOR-Networks, both the filters and the input to convolutional layers are binary. XNOR-Networks approximate convolutions using primarily binary operations. This results in $58\times$ faster convolutional operations (in terms of number of the high precision operations) and $32\times$ memory savings. XNOR-Nets offer the possibility of running state-of-the-art networks on CPUs (rather than GPUs) in real-time. Our binary networks are simple, accurate, efficient, and work on challenging visual tasks. We evaluate our approach on the ImageNet classification task. The classification accuracy with a Binary-Weight-Network version of AlexNet is the same as the full-precision AlexNet. We compare our method with recent network binarization methods, BinaryConnect and BinaryNets, and outperform these methods by large margins on ImageNet, more than $16\%$ in top-1 accuracy. Our code is available at: \href{url}{http://allenai.org/plato/xnornet}. 
\end{abstract}

\section{Introduction}
\begin{figure}[t]
  \includegraphics[width=\textwidth]{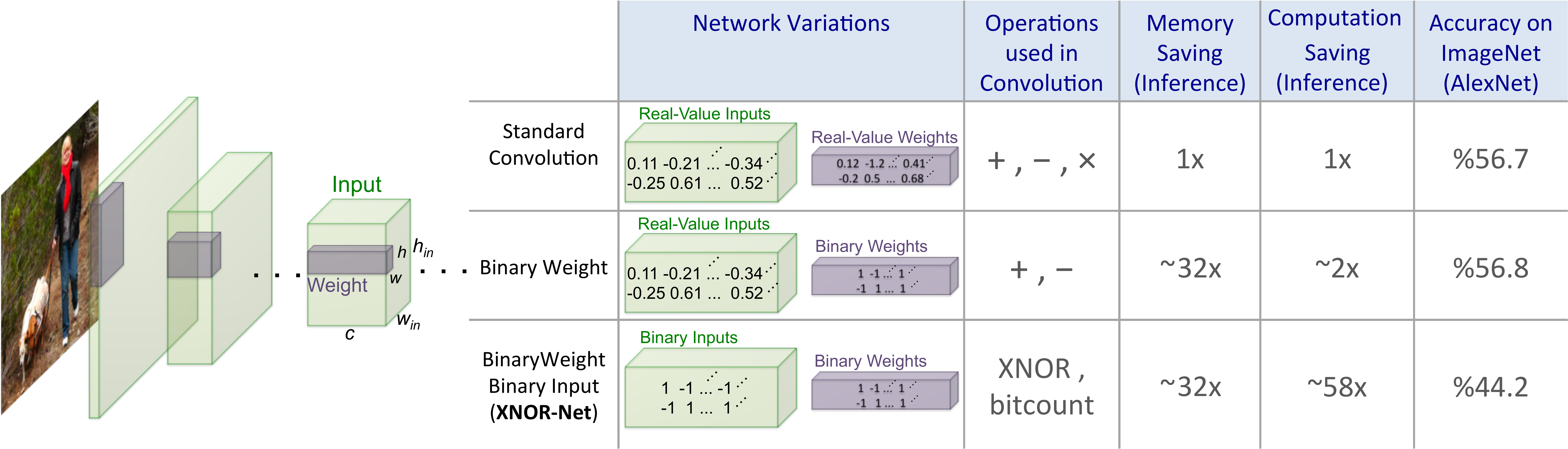}
\caption{We propose two efficient variations of convolutional neural networks. \textbf{Binary-Weight-Networks}, when the weight filters contains binary values. \textbf{XNOR-Networks}, when both weigh and input have binary values. These networks are very efficient in terms of memory and computation, while being very accurate in natural image classification. This offers the possibility of using accurate vision techniques in portable devices with limited resources.}   
\label{fig:teaser}
\end{figure} 

Deep neural networks~(DNN) have shown significant improvements in several application domains including computer vision and speech recognition. In computer vision, a particular type of DNN, known as Convolutional Neural Networks~(CNN), have demonstrated state-of-the-art results in object recognition~\cite{krizhevsky2012imagenet,simonyan2014very,szegedy2015going,kaming2016residual} and detection~\cite{girshick2014rich,girshick2015fast,ren2015faster}.

Convolutional neural networks show reliable results on object recognition and detection that are useful in real world applications. Concurrent to the recent progress in recognition, interesting advancements have been happening in virtual reality (VR by Oculus) \cite{oculus2012oculus}, augmented reality (AR by HoloLens) \cite{gottmer2015merging}, and smart wearable devices. Putting these two pieces together, we argue that it is the right time to equip smart portable devices with the power of state-of-the-art recognition systems. However, CNN-based recognition systems need large amounts of memory and computational power. While they perform well on expensive, GPU-based machines, they are often unsuitable for smaller devices like cell phones and embedded electronics.


For example, AlexNet\cite{krizhevsky2012imagenet} has 61M parameters (249MB of memory) and performs 1.5B high precision operations to classify one image. These numbers are even higher for deeper CNNs \eg VGG \cite{simonyan2014very} (see section \ref{sec:effanalis}). These models quickly overtax the limited storage, battery power, and compute capabilities of smaller devices like cell phones.

In this paper, we introduce simple, efficient, and accurate approximations to CNNs by binarizing the weights and even the intermediate representations in convolutional neural networks. Our binarization method aims at finding the best approximations of the convolutions using binary operations. We demonstrate that our way of binarizing neural networks results in ImageNet classification accuracy numbers that are comparable to standard full precision networks while requiring a significantly less memory and fewer floating point operations.

We study two approximations: Neural networks with binary weights and XNOR-Networks. In \textbf{Binary-Weight-Networks} all the weight values are approximated with binary values. A convolutional neural network with binary weights is significantly smaller ($\sim32\times$) than an equivalent network with single-precision weight values. In addition, when weight values are binary, convolutions can be estimated  by only addition and subtraction (without multiplication), resulting in $\sim2\times$ speed up. Binary-weight approximations of large CNNs can fit into the memory of even small, portable devices while maintaining the same level of accuracy (See Section~\ref{sec:effanalis} and \ref{sec:classification}).

To take this idea further, we introduce \textbf{XNOR-Networks} where both the weights and the inputs to the convolutional and fully connected layers are approximated with binary values\footnote{fully connected layers can be implemented by convolution, therefore, in the rest of the paper, we refer to them also as convolutional layers~\cite{long2015fully}.}. Binary weights and binary inputs allow an efficient way of implementing convolutional operations.  If all of the operands of the convolutions are binary, then the convolutions can be estimated by XNOR and bitcounting operations \cite{courbariaux2016binarynet}. XNOR-Nets result in accurate approximation of CNNs while offering $\sim58\times$ speed up in CPUs (in terms of number of the high precision operations). This means that XNOR-Nets can enable real-time inference in devices with small memory and no GPUs (Inference in XNOR-Nets can be done very efficiently on CPUs).

To the best of our knowledge this paper is  the first attempt to present an evaluation of  binary neural networks on large-scale datasets like ImageNet. Our experimental results show that our proposed method for binarizing convolutional neural networks outperforms the state-of-the-art network binarization method of \cite{courbariaux2016binarynet}  by a large margin ($16.3\%$) on top-1 image classification in the ImageNet challenge ILSVRC2012. Our contribution is two-fold: First, we introduce a new way of binarizing the weight values in convolutional neural networks and show the advantage of our solution compared to state-of-the-art solutions. Second, we introduce XNOR-Nets, a deep neural network model with binary weights and binary inputs and show that XNOR-Nets can obtain similar classification accuracies compared to standard networks while being significantly more efficient. Our code is available at: \href{url}{http://allenai.org/plato/xnornet} 




\section{Related Work}
\label{sec:related}
Deep neural networks often suffer from over-parametrization and large amounts of redundancy in their models. This typically results in inefficient computation and memory usage\cite{denil2013predicting}. Several methods have been proposed to address efficient training and inference in deep neural networks.

\textbf{Shallow networks:} Estimating a deep neural network with a shallower model reduces the size of a network. Early theoretical work by Cybenko shows that a network with a large enough single hidden layer of sigmoid units can approximate any decision boundary \cite{cybenko1989approximation}. In several areas (\eg vision and speech), however, shallow networks cannot compete with deep models \cite{seide2011conversational}. \cite{dauphin2013big} trains a shallow network on SIFT features to classify the ImageNet dataset. They show it is difficult to train shallow networks with large number of parameters. \cite{ba2014deep} provides empirical evidence on small datasets (\eg CIFAR-10) that shallow nets are capable of learning the same functions as deep nets. In order to get the similar accuracy, the number of parameters in the shallow network must be close to the number of parameters in the deep network. They do this by first training a state-of-the-art deep model, and then training a shallow model to mimic the deep model. These methods are different from our approach because we use the standard deep architectures not the shallow estimations.

\textbf{Compressing pre-trained deep networks:} Pruning redundant, non-informative weights in a previously trained network reduces the size of the network at inference time. Weight decay \cite{hanson1989comparing} was an early method for pruning a network. Optimal Brain Damage
\cite{lecun1989optimal} and Optimal Brain Surgeon \cite{hassibi1993second} use the Hessian of the loss function to prune a network by reducing the number of connections. Recently \cite{han2015learning} reduced the number of parameters by an order of magnitude in several state-of-the-art neural networks by pruning. \cite{van2015cross} proposed to reduce the number of activations for compression and acceleration. Deep compression \cite{han2015deep} reduces the storage and energy required to run inference on large networks so they can be deployed on mobile devices. They remove the redundant connections and quantize weights so that multiple connections share the same weight, and then they use Huffman coding to compress the weights. HashedNets \cite{chen2015compressing} uses a hash function to reduce model size by randomly grouping the weights, such that connections in a hash bucket use a single parameter value. Matrix factorization has been used by \cite{denton2014exploiting,jaderberg2014speeding}. We are different from these approaches because we do not use a pretrained network. We train binary networks from scratch.    

\textbf{Designing compact layers:} Designing compact blocks at each layer of a deep network can help to save memory and computational costs. Replacing the fully connected layer with global average pooling was examined in the Network in Network architecture \cite{lin2013network}, GoogLenet\cite{szegedy2015going} and Residual-Net\cite{kaming2016residual}, which achieved state-of-the-art results on several benchmarks. The bottleneck structure in Residual-Net \cite{kaming2016residual} has been proposed to reduce the number of parameters and improve speed. Decomposing $3\times 3$ convolutions with two $1\times 1$ is used in \cite{inception2016Szegedy} and resulted in state-of-the-art performance on object recognition. Replacing $3\times 3$ convolutions with $1\times 1$ convolutions is used in \cite{iandola2016squeezenet} to create a very compact neural network that can achieve $\sim50\times$ reduction in the number of parameters while obtaining high accuracy. Our method is different from this line of work because we use the full network (not the compact version) but with binary parameters.      

\textbf{Quantizing parameters:} High precision parameters are not very important in achieving high performance in deep networks. \cite{gong2014compressing} proposed to quantize the weights of fully connected layers in a deep network by vector quantization techniques. They showed just thresholding the weight values at zero only decreases the top-1 accuracy on ILSVRC2012 by less than $\%$10. \cite{arora2013provable} proposed a provably polynomial time algorithm for training a sparse networks with +1/0/-1 weights. A fixed-point implementation of 8-bit integer was compared with 32-bit floating point activations in \cite{vanhoucke2011improving}. Another fixed-point network with ternary weights and 3-bits activations was presented by \cite{hwang2014fixed}. Quantizing a network with $L_2$ error minimization achieved better accuracy on MNIST and CIFAR-10 datasets in \cite{anwar2015fixed}. \cite{lin2015neural} proposed a back-propagation process by quantizing the representations at each layer of the network. To convert some of the remaining multiplications into binary shifts the neurons get restricted values of power-of-two integers. In \cite{lin2015neural} they carry the full precision weights during the test phase, and only quantize the neurons during the back-propagation process, and not during the forward-propagation. Our work is similar to these methods since we are quantizing the parameters in the network. But our quantization is the extreme scenario +1,-1.      

\textbf{Network binarization:} These works are the most related to our approach. Several methods attempt to binarize the weights and the activations in neural networks.The performance of highly quantized networks (\eg binarized) were believed to be very poor due to the destructive property of binary quantization \cite{courbariaux2014training}. Expectation BackPropagation (EBP) in \cite{soudry2014expectation} showed high performance can be achieved by a network with binary weights and binary activations. This is done by a variational Bayesian approach, that infers networks with binary weights and neurons. A fully binary network at run time presented in \cite{esser2015backpropagation} using a similar approach to EBP, showing significant improvement in energy efficiency. In EBP the binarized parameters were only used during inference. BinaryConnect \cite{courbariaux2015binaryconnect} extended the probablistic idea behind EBP. Similar to our approach, BinaryConnect uses the real-valued version of the weights as a key reference for the binarization process. The real-valued weight updated using the back propagated error by simply ignoring the binarization in the update. BinaryConnect achieved state-of-the-art results on small datasets (\eg CIFAR-10, SVHN). Our experiments shows that this method is not very successful on large-scale datsets (\eg ImageNet). BinaryNet\cite{courbariaux2016binarynet} propose an extention of BinaryConnect, where both weights and activations are binarized. Our method is different from them in the binarization method and the network structure. We also compare our method with BinaryNet on ImageNet, and our method outperforms BinaryNet by a large margin.\cite{wan2013regularization} argued that the noise introduced by weight binarization provides a form of regularization, which could help to improve test accuracy. This method binarizes weights while maintaining full precision activation. \cite{baldassi2015subdominant} proposed fully binary training and testing in an array of committee machines with randomized input. \cite{kim2016bitwise} retraine a previously trained neural network with binary weights and binary inputs.

\section{Binary Convolutional Neural Network}
We represent an $L$-layer CNN architecture with a triplet $\langle \mathcal{I},\mathcal{W},\ast \rangle$. $\mathcal{I}$ is a set of tensors, where each element $\bI=\mathcal{I}_{l(l=1,\dots,L)}$ is the input tensor for the $l^{\text{th}}$ layer of CNN (Green cubes in figure \ref{fig:teaser}). $\mathcal{W}$ is a set of tensors, where each element in this set $\bW =\mathcal{W}_{lk (k=1,\dots,K^{l})}$ is the $k^{\text{th}}$ weight filter in the $l^\text{th}$ layer of the CNN. $K^l$ is the number of weight filters in the $l^\text{th}$ layer of the CNN. $\ast$ represents a convolutional operation with $\bI$ and $\bW$ as its operands\footnote{In this paper we assume convolutional filters do not have bias terms}. $\bI \in \real^{c \times w_{in} \times h_{in}}$, where $(c,w_{in},h_{in})$ represents \textit{channels, width} and \textit{height} respectively.$\bW \in \real^{c\times w\times h}$, where $w \leq w_{in}, ~~h \leq h_{in}$. We propose two variations of binary CNN: \textbf{Binary-weights}, where the elements of $\mathcal{W}$ are binary tensors and \textbf{XNOR-Networks}, where elements of both $\mathcal{I}$ and $\mathcal{W}$ are binary tensors.         

\subsection{Binary-Weight-Networks}
\label{sec:binweight}
In order to constrain a convolutional neural network $\langle \mathcal{I},\mathcal{W},\ast \rangle$ to have binary weights, we estimate the real-value weight filter $\bW \in \mathcal{W}$ using a binary filter $\bB \in \{ +1,-1 \}^{c \times w \times h}$  and a scaling factor $\alpha \in \real^+$ such that $\bW \approx \alpha\bB$. A convolutional operation can be appriximated by: 
\begin{eqnarray}
\label{eq:binconvweight}
\begin{aligned}
  \bI\ast\bW\approx \left(\bI\oplus\bB\right)\alpha
\end{aligned}
\end{eqnarray}
where, $\oplus$ indicates a convolution without any multiplication. Since the weight values are binary, we can implement the convolution with additions and subtractions. The binary weight filters reduce memory usage by a factor of $\sim32\times$ compared to single-precision filters. We represent a CNN with binary weights by $\langle \mathcal{I},\mathcal{B},\mathcal{A},\oplus \rangle$, where $\mathcal{B}$ is a set of binary tensors and $\mathcal{A}$ is a set of positive real scalars, such that $\bB = \mathcal{B}_{lk}$ is a binary filter and $\alpha=\mathcal{A}_{lk}$ is an scaling factor and $\mathcal{W}_{lk} \approx \mathcal{A}_{lk}\mathcal{B}_{lk} $

\subsubsection{Estimating binary weights:}
Without loss of generality we assume $\bW , \bB$ are vectors in $\real^n$, where $n= c \times w \times h$. To find an optimal estimation for $\bW \approx \alpha\bB$, we solve the following optimization:   
\begin{eqnarray}
 \label{eq:binestimate}
 \begin{aligned} 
 J(\bB,\alpha) = \Vert \bW - \alpha\bB \Vert^2 \\
 \alpha^*, \bB^* = \argmin{\alpha,\bB} J(\bB,\alpha)
 \end{aligned} 
\end{eqnarray}

by expanding equation \ref{eq:binestimate}, we have 
\begin{eqnarray}
 \label{eq:expand}
  J(\bB,\alpha) = \alpha^2\trans{\bB}\bB - 2\alpha\trans{\bW}\bB+\trans{\bW}\bW
\end{eqnarray}

since $\bB \in \{+1,-1\}^n$, $\trans{\bB}\bB=n$ is a constant .  $\trans{\bW}\bW$ is also a constant because $\bW$ is a known variable. Lets define $\bc=\trans{\bW}\bW$. Now, we can rewrite the equation \ref{eq:expand} as follow: $J(\bB,\alpha) =\alpha^2 n - 2\alpha\trans{\bW}\bB+\bc$. The optimal solution for $\bB$ can be achieved by maximizing the following constrained optimization: (note that $\alpha$ is a positive value in equation \ref{eq:binestimate}, therefore it can be ignored in the maximization)
\begin{eqnarray}
 \label{eq:optb}
 \bB^* = \argmax{\bB}\{\trans{\bW}\bB\} ~~~ s.t.~~ \bB \in \{+1,-1\}^n
\end{eqnarray}

This optimization can be solved by assigning $\bB_i = +1$ if $\bW_i \geq 0$ and $\bB_i = -1$ if $\bW_i < 0$, therefore the optimal solution is $\bB^* = \sign(\bW)$. In order to find the optimal value for the scaling factor $\alpha^*$, we take the derivative of $J$ with respect to $\alpha$ and set it to zero: 
\begin{eqnarray}
 \label{eq:optalpha}
 \alpha^* = \frac{\trans{\bW}\bB^*}{n}
\end{eqnarray}

By replacing $\bB^*$ with $\sign(\bW)$
\begin{eqnarray}
 \label{eq:optalpha2}
 \alpha^* = \frac{\trans{\bW}\sign(\bW)}{n} = \frac{\sum \vert\bW_i\vert}{n} = \frac{1}{n}\Vert\bW\Vert_{\ell 1}
\end{eqnarray}
therefore, the optimal estimation of a binary weight filter can be simply achieved by taking the sign of weight values. The optimal scaling factor is the average of absolute weight values.

\subsubsection{Training Binary-Weights-Networks:}
\label{binweight}

\renewcommand{\algorithmicrequire}{\textbf{Input:}}
\renewcommand{\algorithmicensure}{\textbf{Output:}}
\begin{algorithm}[thb]		
{\footnotesize
  \caption{Training an $L$-layers CNN with binary weights:}
  \label{alg:trainbinconv}       
  \begin{algorithmic}[1]
  \REQUIRE A minibatch of inputs and targets  ($\bI, \bY$), cost function $C(\bY,\hat{\bY})$, current weight $\mathcal{W}^t$ and current learning rate $\eta^t$.     
  \ENSURE updated weight $\mathcal{W}^{t+1}$ and updated learning rate $\eta^{t+1}$. 
  \STATE Binarizing weight filters:
  \FOR{$l=1$ to $L$}
      \FOR{$k^{\text{th}}$ filter in $l^{\text{th}}$ layer}
          \STATE $\mathcal{A}_{lk}=\frac{1}{n}\Vert\mathcal{W}_{lk}^t\Vert_{\ell 1}$
          \STATE $\mathcal{B}_{lk}=\sign(\mathcal{W}_{lk}^t)$
          \STATE $\widetilde{\mathcal{W}}_{lk}= \mathcal{A}_{lk}\mathcal{B}_{lk}$
      \ENDFOR
   \ENDFOR
  \STATE $\hat{\bY}=$ ~~\textbf{BinaryForward}$(\bI,\mathcal{B},\mathcal{A})$~~//~{\scriptsize standard forward propagation except that convolutions are computed using equation \ref{eq:binconvweight} or \ref{eq:binconve} }
  
  \STATE $\frac{\partial C}{\partial \widetilde{\mathcal{W}}} =$ \textbf{BinaryBackward}$(\frac{\partial C}{\partial \hat{\bY}}, \widetilde{\mathcal{W}})$ ~~//~{\scriptsize standard backward propagation except that gradients are computed using $\widetilde{\mathcal{W}}$ instead of $\mathcal{W}^t$}  
 \STATE $\mathcal{W}^{t+1}=$ \textbf{UpdateParameters}$(\mathcal{W}^t,\frac{\partial C}{\partial \widetilde{\mathcal{W}}},\eta_t)$~~//~{\scriptsize Any update rules (\eg SGD or ADAM)} 
 \STATE $\eta^{t+1}=$ \textbf{UpdateLearningrate}$(\eta^t, t)$~~//~ {\scriptsize Any learning rate scheduling function}
  \end{algorithmic}
  } \normalsize
\end{algorithm}

Each iteration of training a CNN involves three steps; forward pass, backward pass and parameters update. To train a CNN with binary weights (in convolutional layers), we only binarize the weights during the forward pass and backward propagation. To compute the gradient for $\sign$ function $\sign(r)$, we follow the same approach as \cite{courbariaux2016binarynet}, where $\frac{\partial \sign}{\partial r}=r1_{\vert r\vert\leq1}$. The gradient in backward after the scaled sign function is $\frac{\partial{C}}{\partial{W_i}}= \frac{\partial{C}}{\widetilde{W_i}}(\frac{1}{n}+\frac{\partial \sign}{\partial W_i}\alpha)$. For updating the parameters, we use the high precision (real-value) weights. Because, in gradient descend the parameter changes are tiny, binarization after updating the parameters ignores these changes and the training objective can not be improved. \cite{courbariaux2016binarynet,courbariaux2015binaryconnect} also employed this strategy to train a binary network.

Algorithm \ref{alg:trainbinconv} demonstrates our procedure for training a CNN with binary weights. First, we binarize the weight filters at each layer by computing $\mathcal{B}$ and $\mathcal{A}$. Then we call forward propagation using binary weights and its corresponding scaling factors, where all the convolutional operations are carried out by equation \ref{eq:binconvweight}. Then, we call backward propagation, where the gradients are computed with respect to the estimated weight filters $\widetilde{\mathcal{W}}$. Lastly, the parameters and the learning rate gets updated by an update rule \eg SGD update with momentum or ADAM \cite{kingma2014adam}.   

Once the training finished, there is no need to keep the real-value weights. Because, at inference we only perform forward propagation with the binarized weights. 

\subsection{XNOR-Networks}
\label{sec:bininput}

\begin{figure}[t]
\centering
  \includegraphics[width=\textwidth]{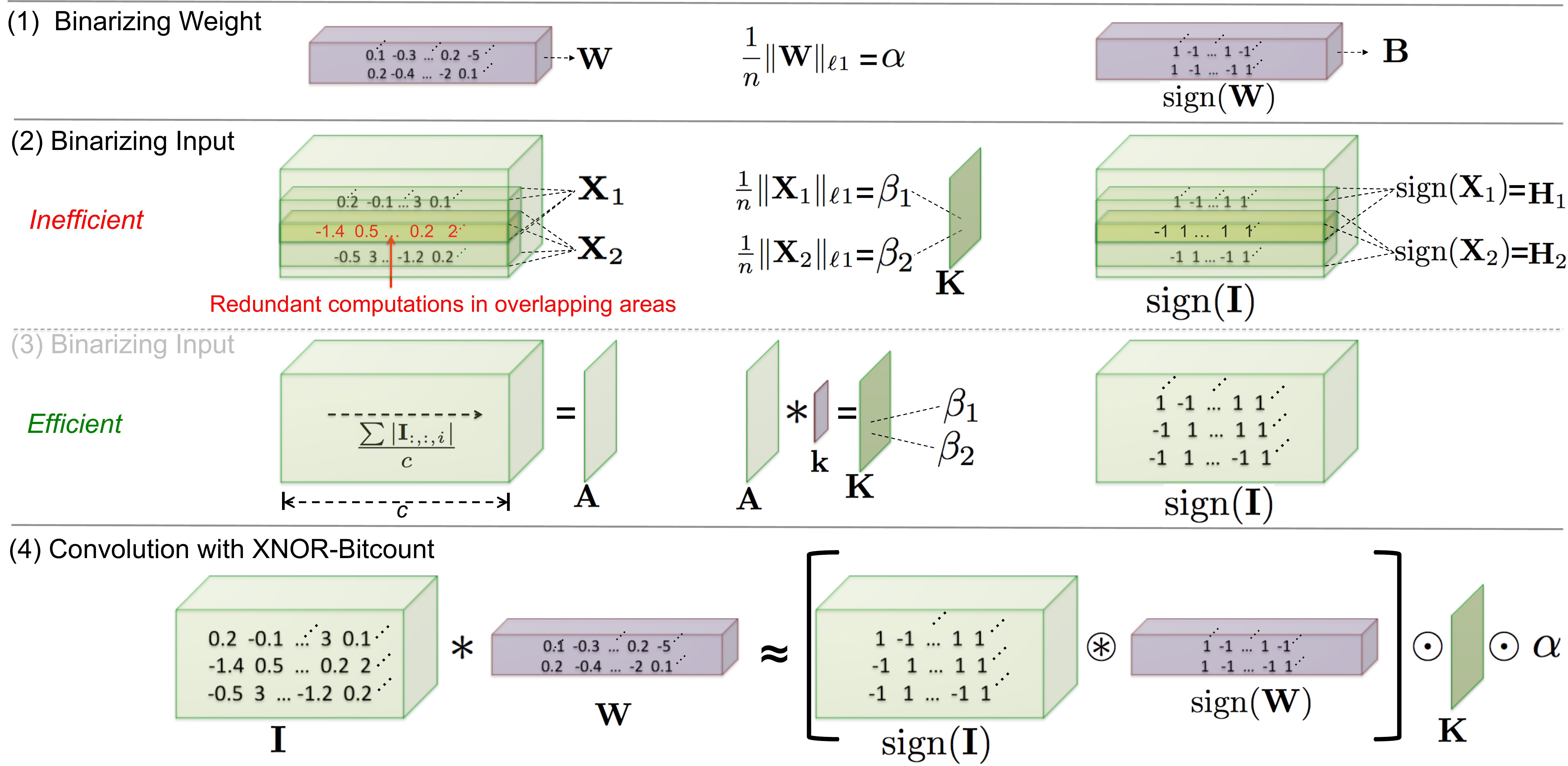}
\caption{\footnotesize This figure illustrates the procedure explained in section~\ref{sec:binconv} for approximating a convolution using binary operations. }
\label{fig:binconv}
\end{figure} 

So far, we managed to find binary weights and a scaling factor to estimate the real-value weights. The inputs to the convolutional layers are still real-value tensors. Now, we explain how to binarize both weigths and inputs, so convolutions can be implemented efficiently using XNOR and bitcounting operations. This is the key element of our XNOR-Networks.      
In order to constrain a convolutional neural network $\langle \mathcal{I},\mathcal{W},\ast \rangle$ to have binary weights and binary inputs, we need to enforce binary operands at each step of the convolutional operation. A convolution consist of repeating a shift operation and a dot product. Shift operation moves the weight filter over the input and the dot product performs element-wise multiplications between the values of the weight filter and the corresponding part of the input. If we express dot product in terms of binary operations, convolution can be approximated using binary operations. Dot product between two binary vectors can be implemented by XNOR-Bitcounting operations \cite{courbariaux2016binarynet}.  In this section, we explain how to approximate the dot product between two vectors in $\real^n$ by a dot product between two vectors in $\{+1,-1\}^n$. Next, we demonstrate how to use this approximation for estimating a convolutional operation between two tensors.

\subsubsection{Binary Dot Product:}
  To approximate the dot product between $\bX,\bW \in \real^n$ such that  $\trans{\bX}\bW \approx \beta\trans{\bH}\alpha\bB$, where $\bH, \bB \in \{+1,-1\}^n$ and $\beta,\alpha \in \real^+$, we solve the following optimization: 
\begin{eqnarray}
 \label{eq:dotbinestimate}
 \alpha^*, \bB^*, \beta^*, \bH* = \argmin{\alpha,\bB,\beta,\bH} \Vert \bX\odot\bW - \beta\alpha\bH\odot\bB \Vert
\end{eqnarray}
where $\odot$ indicates element-wise product. We define $\bY \in \real^n$ such that $\bY_i= \bX_i\bW_i$, $\bC \in \{+1,-1\}^n$ such that $\bC_i = \bH_i\bB_i$ and $\gamma \in \real^+$ such that $\gamma = \beta\alpha$. The equation \ref{eq:dotbinestimate} can be written as: \begin{eqnarray}
 \label{eq:dotbinestimaterewrite}
 \gamma^*, \bC^* = \argmin{\gamma,\bC} \Vert \bY - \gamma\bC \Vert
\end{eqnarray}
 the optimal solutions can be achieved from equation \ref{eq:binestimate} as follow
\begin{eqnarray}
\label{eq:dotbinoptsol}
\begin{aligned}
  \bC^* = \sign(\bY) = \sign(\bX)\odot\sign(\bW) = \bH^*\odot\bB^*\\
\end{aligned}
\end{eqnarray}
Since $\vert\bX_i\vert,\vert\bW_i\vert$ are independent, knowing that $\bY_i= \bX_i\bW_i$ then, \\    $\bE\left[\vert\bY_i\vert\right]=\bE\left[\vert\bX_i\vert\vert\bW_i\vert\right]=\bE\left[\vert\bX_i\vert\right]\bE\left[\vert\bW_i\vert\right]$ therefore, 
\begin{eqnarray}
\label{eq:gammaopt}
\begin{aligned}
  \gamma^* = \frac{\sum \vert\bY_i\vert}{n} = \frac{\sum \vert\bX_i\vert\vert\bW_i\vert}{n} \approx \left(\frac{1}{n}\Vert\bX\Vert_{\ell 1}\right)\left(\frac{1}{n}\Vert\bW\Vert_{\ell 1}\right) = \beta^*\alpha^*  
\end{aligned}
\end{eqnarray}
\subsubsection{Binary Convolution:}
\label{sec:binconv}
Convolving weight filter $\bW \in \real^{c \times w \times h}$ (where $w_{in} \gg w,~~ h_{in}\gg h$) with the input tensor $\bI \in \real^{c \times w_{in} \times h_{in}}$ requires computing the scaling factor $\beta$ for all possible sub-tensors in $\bI$ with same size as $\bW$. Two of these sub-tensors are illustrated in figure~\ref{fig:binconv} (second row) by $\bX_1$ and $\bX_2$. Due to overlaps between subtensors, computing $\beta$ for all possible sub-tensors leads to a large number of redundant computations. To overcome this redundancy, first, we compute a matrix $\bA = \frac{\sum\vert\bI_{:,:,i}\vert}{c}$, which is the average over absolute values of the elements in the input $\bI$ across the channel. Then we convolve $\bA$ with a 2D filter $\bk \in \real^{w \times h}$, $\bK = \bA \ast \bk$,  where $\forall ij~~ \bk_{ij}=\frac{1}{w\times h}$. $\bK$ contains scaling factors $\beta$ for all sub-tensors in the input $\bI$. $\bK_{ij}$ corresponds to $\beta$ for a sub-tensor centered at the location $ij$ (across width and height). This procedure is shown in the third row of figure~\ref{fig:binconv}. Once we obtained the scaling factor $\alpha$ for the weight and $\beta$ for all sub-tensors in $\bI$ (denoted by $\bK$), we can approximate the convolution between input $\bI$ and weight filter $\bW$ mainly using binary operations: 
\begin{eqnarray}
\label{eq:binconve}
\begin{aligned}
  \bI\ast\bW\approx \left(\sign(\bI)\circledast\sign(\bW)\right)\odot\bK\alpha
\end{aligned}
\end{eqnarray}
where $\circledast$ indicates a convolutional operation using XNOR and bitcount operations. This is illustrated in the last row in figure~\ref{fig:binconv}. Note that the number of non-binary operations is very small compared to binary operations.  

\subsubsection{Training XNOR-Networks:}

\begin{figure}[t]
\centering
  \includegraphics[height=0.16\textwidth]{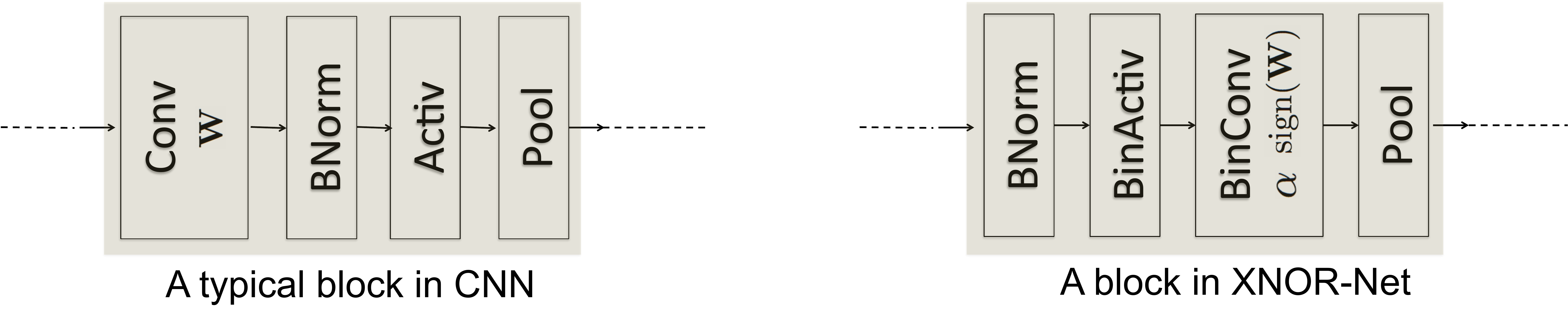}
\caption{\footnotesize This figure contrasts the block structure in our XNOR-Network (right) with a typical CNN (left). }
\label{fig:binblock}
\end{figure} 

A typical block in CNN contains several different layers. Figure~\ref{fig:binblock} (left) illustrates a typical block in a CNN. This block has four layers in the following order: 1-Convolutional, 2-Batch Normalization, 3-Activation and 4-Pooling. Batch Normalization layer\cite{ioffe2015batch} normalizes the input batch by its mean and variance. The activation is an element-wise non-linear function (\eg Sigmoid, ReLU). The pooling layer applies any type of pooling (\eg max,min or average) on the input batch. 
Applying pooling on binary input results in significant loss of information. For example, max-pooling on binary input returns a tensor that most of its elements are equal to $+1$. Therefore, we put the pooling layer after the convolution. To further decrease the information loss due to binarization, we normalize the input before binarization. This ensures the data to hold zero mean, therefore, thresholding at zero leads to less quantization error. The order of layers in a block of binary CNN is shown in Figure~\ref{fig:binblock}(right). 

The binary activation layer(BinActiv) computes $\bK$ and $\sign(\bI)$ as explained in section~\ref{sec:binconv}. In the next layer (BinConv), given $\bK$ and $\sign(\bI)$, we compute binary convolution by equation~\ref{eq:binconve}. Then at the last layer (Pool), we apply the pooling operations. We can insert a non-binary activation(\eg ReLU) after binary convolution. This helps when we use state-of-the-art networks (\eg AlexNet or VGG).   

Once we have the binary CNN structure, the training algorithm would be the same as algorithm~\ref{alg:trainbinconv}. 

\textbf{Binary Gradient:} The computational bottleneck in the backward pass at each layer is computing a convolution between weight filters($w$) and the gradients with respect of the inputs ($g^{in}$). Similar to binarization in the forward pass, we can binarize $g^{in}$ in the backward pass. This leads to a very efficient training procedure using binary operations. Note that if we use equation \ref{eq:optalpha2} to compute the scaling factor for $g^{in}$, the direction of maximum change for SGD would be diminished. To preserve the maximum change in all dimensions, we use $\max_i(|g^{in}_i|)$ as the scaling factor.   

\textbf{$k$-bit Quantization:} So far, we showed 1-bit quantization of weights and inputs using $\sign(x)$ function. One can easily extend the quantization level to $k$-bits by using $q_k(x)=2(\frac{[(2^k-1)(\frac{x+1}{2})]}{2^k-1}-\frac{1}{2})$ instead of the $\sign$ function. Where $[.]$ indicates rounding operation and $x \in [-1,1]$.
 
\section{Experiments}
\label{sec:exp}
We evaluate our method by analyzing its efficiency and accuracy. We measure the efficiency by computing the computational speedup (in terms of number of high precision operation) achieved by our binary convolution vs. standard convolution. To measure accuracy, we perform image classification on the large-scale ImageNet dataset. This paper is the first work that evaluates binary neural networks on the ImageNet dataset. Our binarization technique is general, we can use any CNN architecture. We evaluate AlexNet~\cite{krizhevsky2012imagenet} and two deeper architectures in our experiments. We compare our method with two recent works on binarizing neural networks; BinaryConnect~\cite{courbariaux2015binaryconnect} and BinaryNet~\cite{courbariaux2016binarynet}. The classification accuracy of our binary-weight-network version of AlexNet is as accurate as the full precision version of AlexNet. This classification accuracy outperforms competitors on binary neural networks by a large margin. We also present an ablation study, where we evaluate the key elements of our proposed method; computing scaling factors and our block structure for binary CNN. We shows that our method of computing the scaling factors is important to reach high accuracy.

\subsection{Efficiency Analysis}
\label{sec:effanalis}
In an standard convolution, the total number of operations is $c N_{\bW} N_{\bI}$, where $c$ is the number of channels, $N_{\bW}=wh$ and $N_{\bI}=w_{in}h_{in}$. Note that some modern CPUs can fuse the multiplication and addition as a single cycle operation. On those CPUs, Binary-Weight-Networks does not deliver speed up. Our binary approximation of convolution (equation \ref{eq:binconve}) has  $c N_{\bW}N_{\bI}$ binary operations and $N_{\bI}$ non-binary operations. With the current generation of CPUs, we can perform 64 binary operations in one clock of CPU, therefore the speedup can be computed by $S = \frac{cN_{\bW}N_{\bI}}{\frac{1}{64}cN_{\bW}N_{\bI}+N_{\bI}} = \frac{64cN_{\bW}}{cN_{\bW}+64}$.  

\begin{figure*}[t!]
\centering
\subfloat[]{\includegraphics[width=0.4\textwidth]{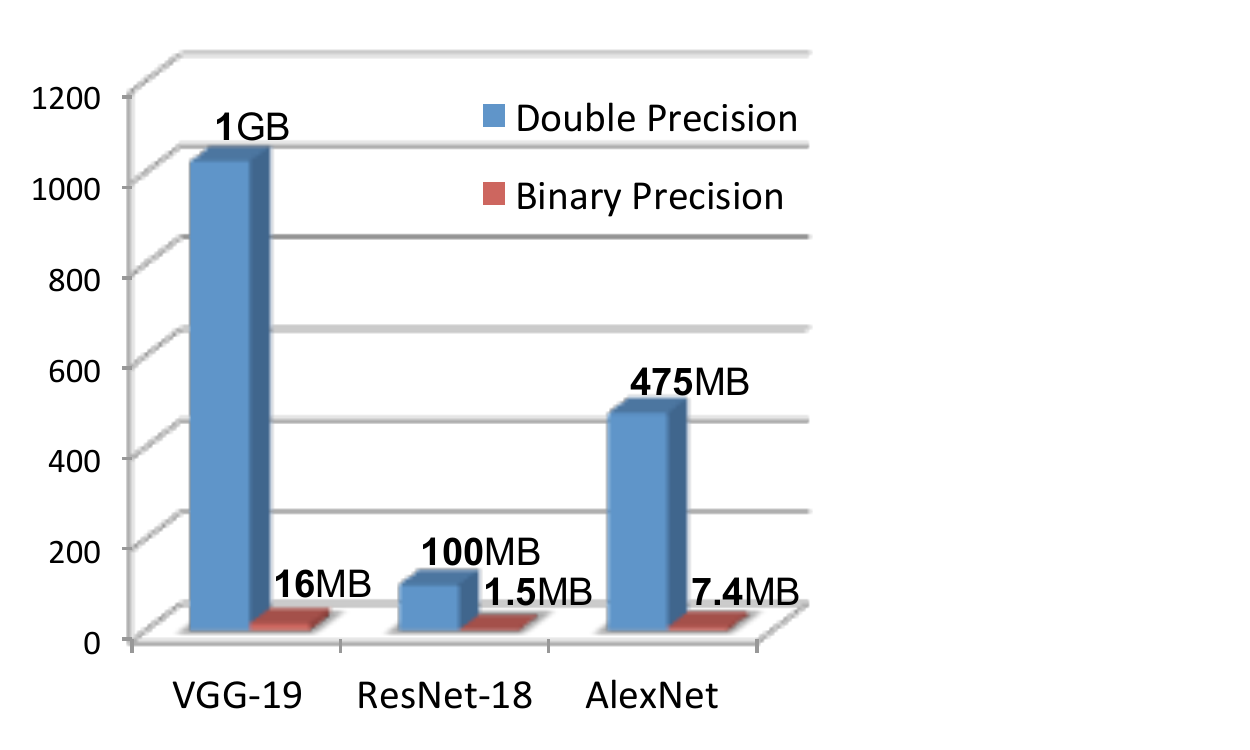}}
\subfloat[]{\includegraphics[height=0.23\textwidth]{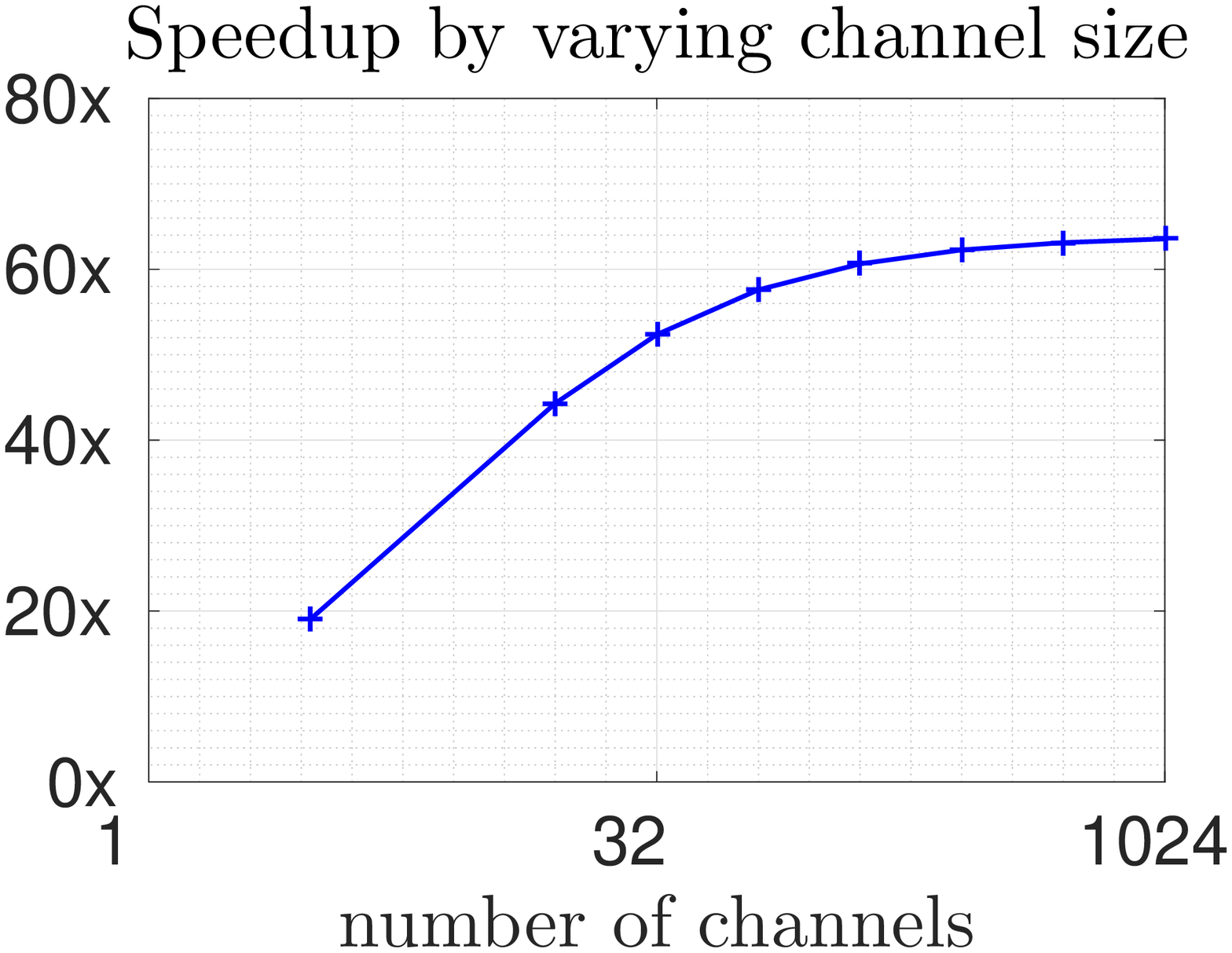}}
\subfloat[]{\includegraphics[height=0.23\textwidth]{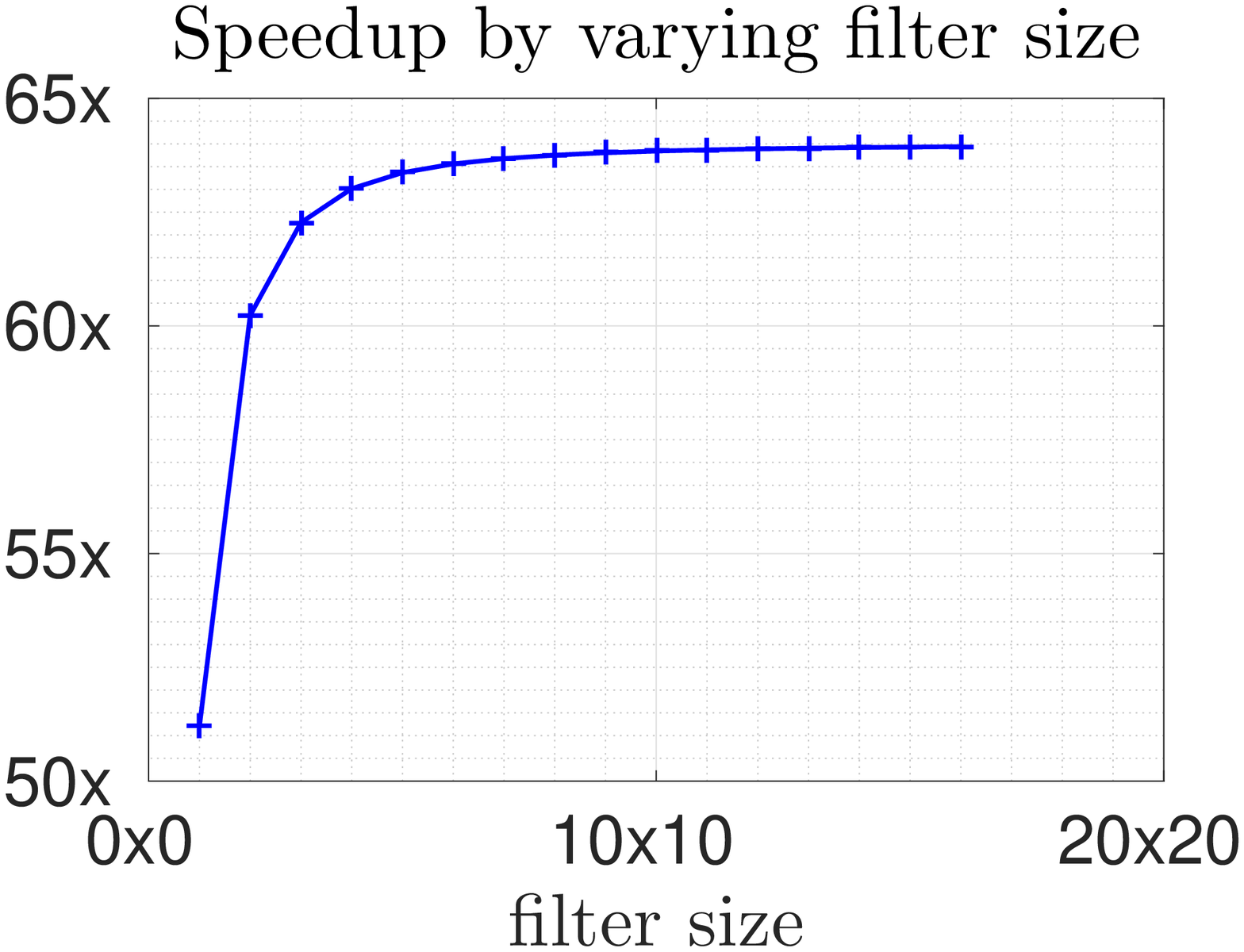}}
\caption{\footnotesize This figure shows the efficiency of binary convolutions in terms of memory(a) and computation(b-c). (a) is contrasting the required memory for binary and double precision weights in three different architectures(AlexNet, ResNet-18 and VGG-19). (b,c) Show speedup gained by binary convolution under (b)-different number of channels and  (c)-different filter size}
\label{fig:eff}
\end{figure*}
The speedup depends on the channel size and filter size but not the input size. In figure \ref{fig:eff}-(b-c) we illustrate the speedup achieved by changing the number of channels and filter size. While changing one parameter, we fix other parameters as follows: $c=256$, $n_{\bI}=14^2$ and $n_{\bW}=3^2$ (majority of convolutions in ResNet\cite{kaming2016residual} architecture have this structure). Using our approximation of convolution we gain 62.27$\times$ theoretical speed up, but in our CPU implementation with all of the overheads, we achieve 58x speed up in one convolution (Excluding the process for memory allocation and memory access). With the small channel size ($c=3$) and filter size ($N_{\bW}=1\times 1$) the speedup is not considerably high. This motivates us to avoid binarization at the first and last layer of a CNN. In the first layer the chanel size is $3$ and in the last layer the filter size is $1\times 1$. A similar strategy was used in ~\cite{courbariaux2016binarynet}. Figure~\ref{fig:eff}-a shows the required memory for three different CNN architectures(AlexNet, VGG-19, ResNet-18) with binary and double precision weights. Binary-weight-networks are so small that can be easily fitted into portable devices. BinaryNet~\cite{courbariaux2016binarynet} is in the same order of memory and computation efficiency as our method. In Figure~\ref{fig:eff}, we show an analysis of computation and memory cost for a binary convolution. The same analysis is valid for BinaryNet and BinaryConnect. The key difference of our method is using a scaling-factor, which does not change the order of efficiency while providing a significant improvement in accuracy.     

\subsection{Image Classification}
\label{sec:classification}
We evaluate the performance of our proposed approach on the task of natural image classification. So far, in the literature, binary neural network methods have presented their evaluations on either limited domain or simplified datasets \eg CIFAR-10, MNIST, SVHN. To compare with state-of-the-art vision, we evaluate our method on ImageNet (ILSVRC2012). ImageNet has $\sim$1.2M train images from 1K categories and 50K validation images. The images in this dataset are natural images with reasonably high resolution compared to the CIFAR and MNIST dataset, which have relatively small images.  We report our classification performance using Top-1 and Top-5 accuracies. We adopt three different CNN architectures as our base architectures for binarization: AlexNet~\cite{krizhevsky2012imagenet}, Residual Networks (known as ResNet)~\cite{kaming2016residual}, and a variant of GoogLenet \cite{szegedy2015going}.We compare our Binary-weight-network (\textbf{BWN}) with BinaryConnect(\textbf{BC})~\cite{courbariaux2015binaryconnect} and our XNOR-Networks(\textbf{XNOR-Net}) with BinaryNeuralNet(\textbf{BNN})~\cite{courbariaux2016binarynet}.
BinaryConnect(BC) is a method for training a deep neural network with binary weights during forward and backward propagations. Similar to our approach, they keep the real-value weights during the updating parameters step. Our binarization is different from BC. The binarization in BC can be either deterministic or stochastic. We use the deterministic binarization for BC in our comparisons because the stochastic binarization is not efficient. The same evaluation settings have been used and discussed in ~\cite{courbariaux2016binarynet}. BinaryNeuralNet(BNN) ~\cite{courbariaux2016binarynet} is a neural network with binary weights and activations during inference and gradient computation in training. In concept, this is a similar approach to our XNOR-Network but the binarization method and the network structure in BNN is different from ours. Their training algorithm is similar to BC and they used deterministic binarization in their evaluations.

\textbf{CIFAR-10 :}
BC and BNN showed near state-of-the-art performance on CIFAR-10, MNIST, and SVHN dataset. BWN and XNOR-Net on CIFAR-10 using the same network architecture as BC and BNN achieve the error rate of 9.88\% and 10.17\% respectively.
In this paper we explore the possibility of obtaining near state-of-the-art results on a much larger and more challenging dataset (ImageNet).

\textbf{AlexNet: }
\begin{figure*}[t]
\centering
\begin{tabular}{cccc}
\includegraphics[height=0.2\textwidth]{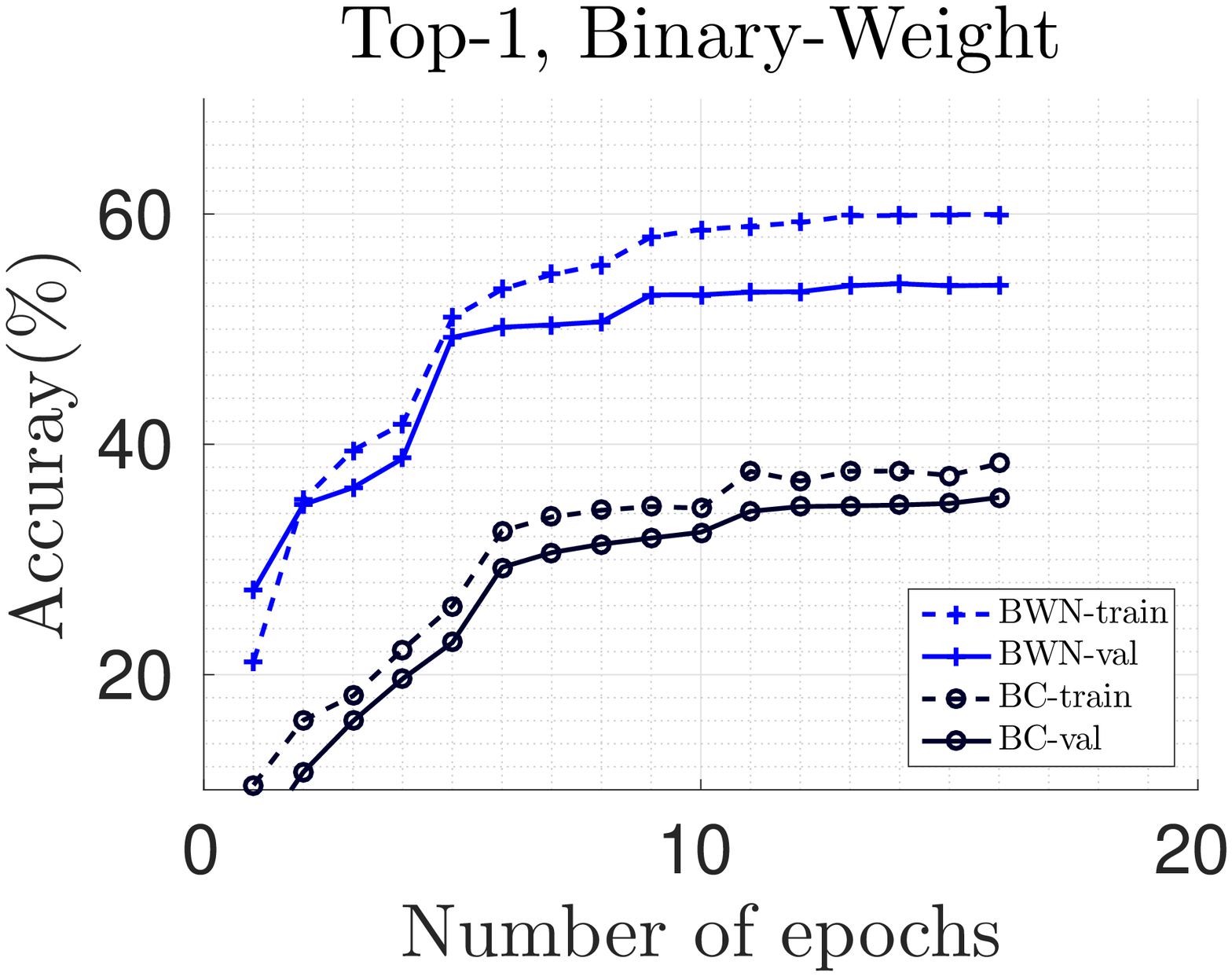}&
\includegraphics[height=0.2\textwidth]{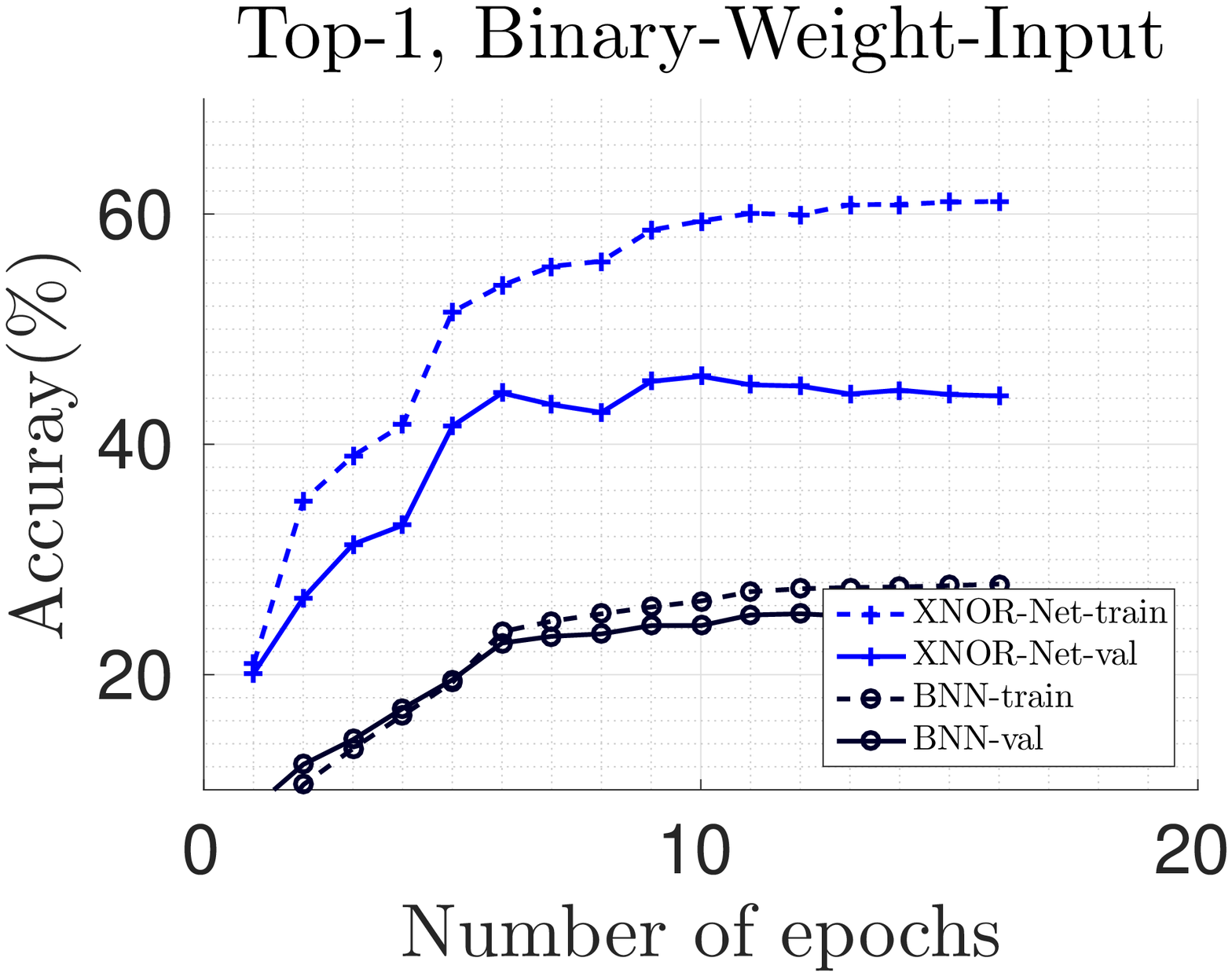}
\includegraphics[height=0.2\textwidth]{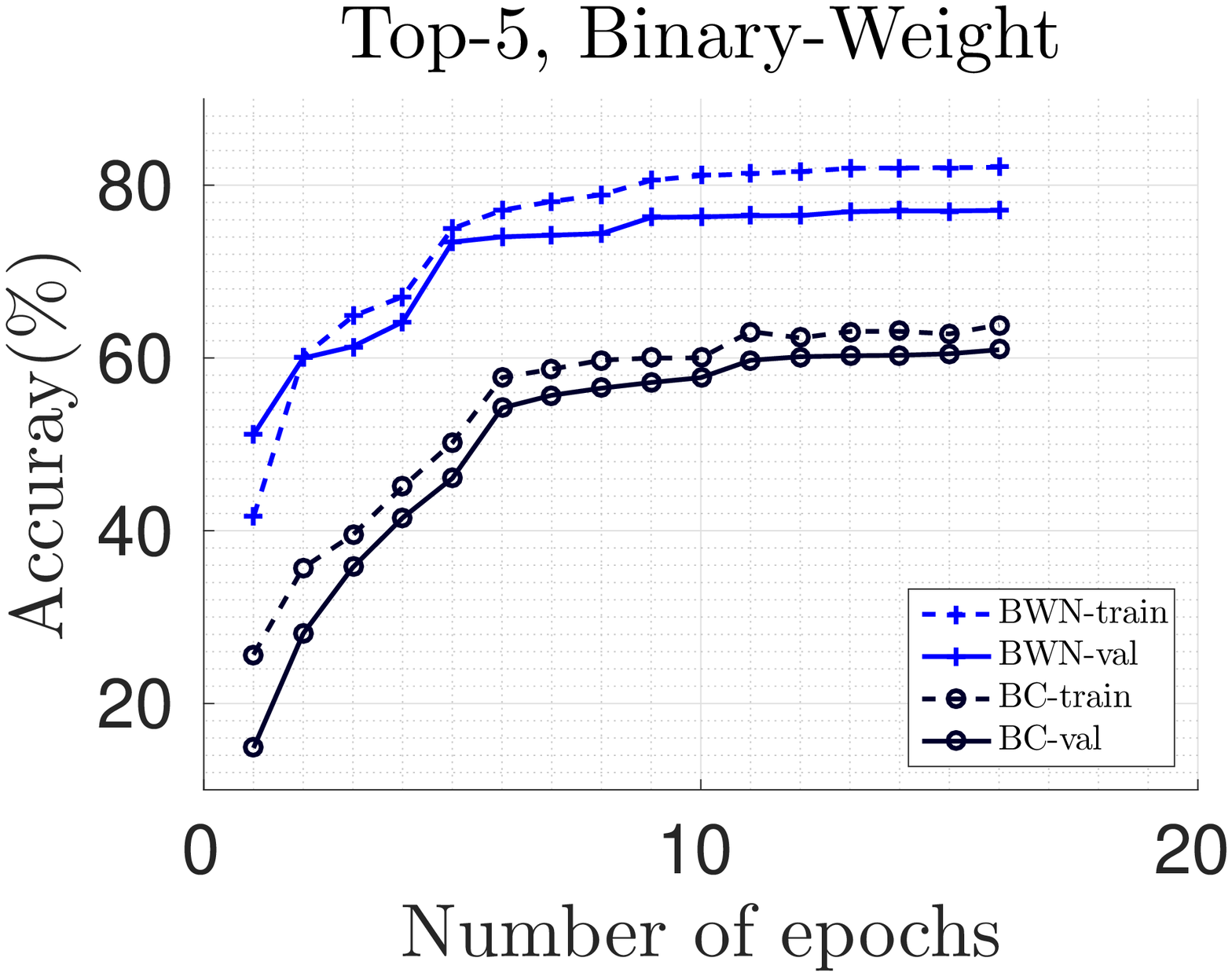}&
\includegraphics[height=0.2\textwidth]{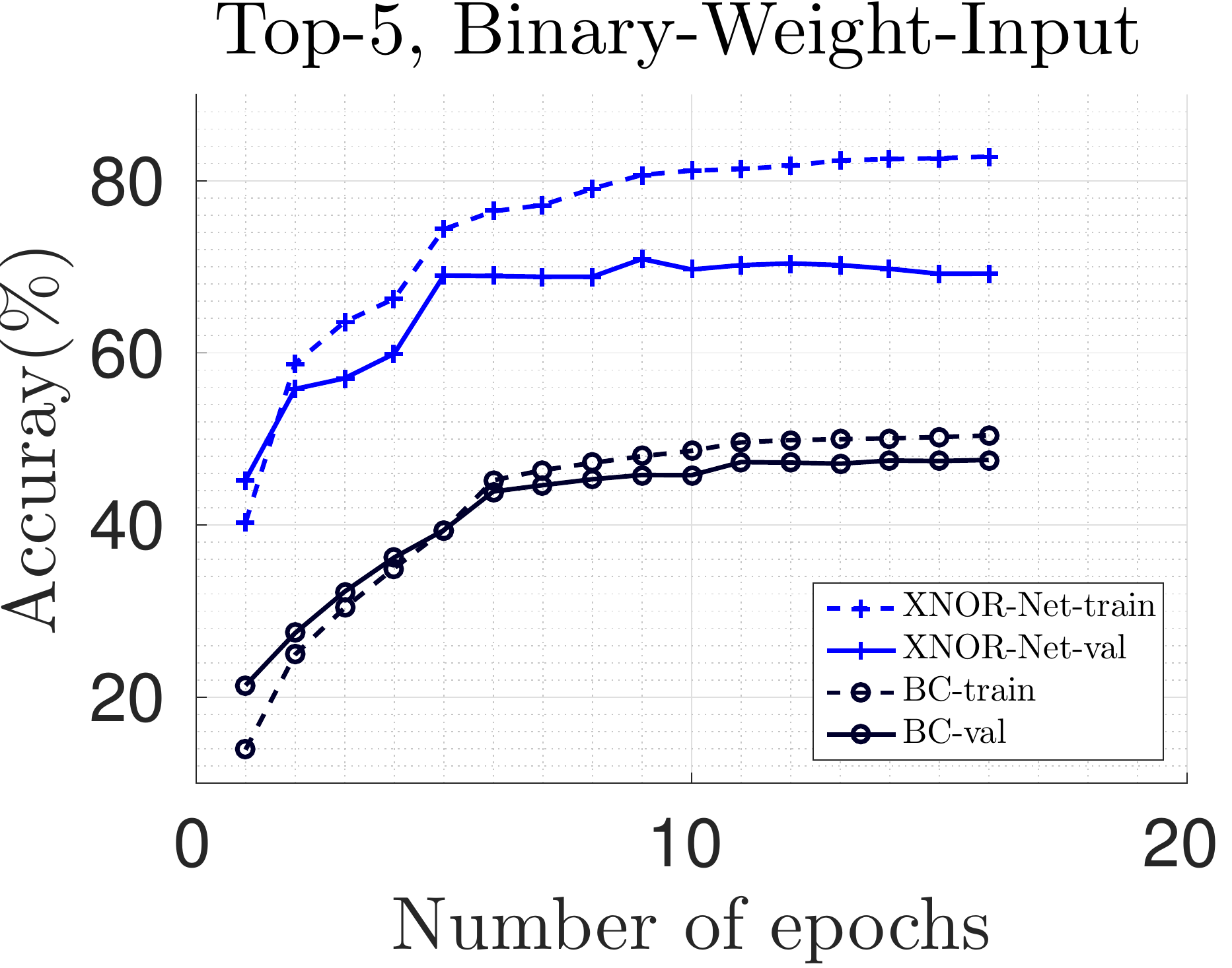}
\end{tabular}
\caption{\footnotesize This figure compares the imagenet classification accuracy on Top-1 and Top-5 across training epochs. Our approaches BWN and XNOR-Net outperform BinaryConnect(BC) and BinaryNet(BNN) in all the epochs by large margin($\sim$17\%).}
\label{fig:alexnetcurve}
\end{figure*}
~\cite{krizhevsky2012imagenet} is a CNN architecture with 5 convolutional layers and two fully-connected layers. This architecture was the first CNN architecture that showed to be successful on ImageNet classification task. This network has 61M parameters. We use AlexNet coupled with batch normalization layers~\cite{ioffe2015batch}.  
\begin{table}
\centering
\begin{tabular}{| c | c | c  | c  || c | c | c  | c  || c | c |}

 \hline
 \multicolumn{10}{|c|}{Classification Accuracy($\%$)} \\
 \hline
  \multicolumn{4}{|c|}{Binary-Weight}& \multicolumn{4}{||c||}{Binary-Input-Binary-Weight}& \multicolumn{2}{|c|}{Full-Precision}\\
  \hline
  \multicolumn{2}{|c|}{BWN}& \multicolumn{2}{|c|}{BC\cite{courbariaux2016binarynet}}& \multicolumn{2}{||c|}{XNOR-Net}& \multicolumn{2}{|c||}{BNN\cite{courbariaux2016binarynet}}&  \multicolumn{2}{c|}{AlexNet\cite{krizhevsky2012imagenet}}\\
  \hline
  \multicolumn{1}{|c|}{Top-1}& \multicolumn{1}{c}{Top-5}& \multicolumn{1}{|c|}{Top-1}& \multicolumn{1}{|c}{Top-5}& \multicolumn{1}{||c|}{Top-1}& \multicolumn{1}{c|}{Top-5}& \multicolumn{1}{|c}{Top-1}& \multicolumn{1}{|c||}{Top-5}& \multicolumn{1}{c|}{Top-1}& \multicolumn{1}{c|}{Top-5}\\
 \hline
 \multicolumn{1}{|c|}{\textbf{56.8}}& \multicolumn{1}{c}{\textbf{79.4}}& \multicolumn{1}{|c|}{35.4}& \multicolumn{1}{|c}{61.0}& \multicolumn{1}{||c|}{\textbf{44.2}}& \multicolumn{1}{c|}{\textbf{69.2}}& \multicolumn{1}{|c}{27.9}& \multicolumn{1}{|c||}{50.42}& \multicolumn{1}{c|}{56.6}& \multicolumn{1}{c|}{80.2}\\
 \hline
\end{tabular}
\caption{\footnotesize This table compares the final accuracies (Top1 - Top5) of the full precision network with our binary precision networks; Binary-Weight-Networks(BWN) and XNOR-Networks(XNOR-Net) and the competitor methods; BinaryConnect(BC) and BinaryNet(BNN).}
\label{table:alexnetcomp}
\end{table}

\textit{Train:}  In each iteration of training, images are resized to have 256 pixel at their smaller dimension and then a random crop of $224\times 224$ is selected for training. We run the training algorithm for 16 epochs with batche size equal to 512. We use negative-log-likelihood over the soft-max of the outputs as our classification loss function. In our implementation of AlexNet we do not use the Local-Response-Normalization(LRN) layer\footnote{Our implementation is followed by https://gist.github.com/szagoruyko/dd032c529048492630fc}. We use SGD with momentum=0.9 for updating parameters in BWN and BC. For XNOR-Net and BNN we used ADAM ~\cite{kingma2014adam}. ADAM converges faster and usually achieves better accuracy for binary inputs~\cite{courbariaux2016binarynet}. The learning rate starts at 0.1 and we apply a learning-rate-decay=0.01 every 4 epochs.    

\textit{Test:} At inference time, we use the $224\times 224$ center crop for forward propagation. 

Figure~\ref{fig:alexnetcurve} demonstrates the classification accuracy for training and inference along the training epochs for top-1 and top-5 scores. The dashed lines represent training accuracy and solid lines shows the validation accuracy. In all of the epochs our method outperforms BC and BNN by large margin ($\sim$17\%). Table~\ref{table:alexnetcomp} compares our final accuracy with BC and BNN. We found that the scaling factors for the weights ($\alpha$) is much more effective than the scaling factors for the inputs ($\beta$). Removing $\beta$ reduces the accuracy by a small margin (less than $1\%$ top-1 alexnet).

\textit{Binary Gradient:} Using XNOR-Net with binary gradient the accuracy of top-1 will drop only by 1.4\%. 
\begin{figure*}[t!]
\centering
\subfloat[]{\includegraphics[height=0.25\textwidth]{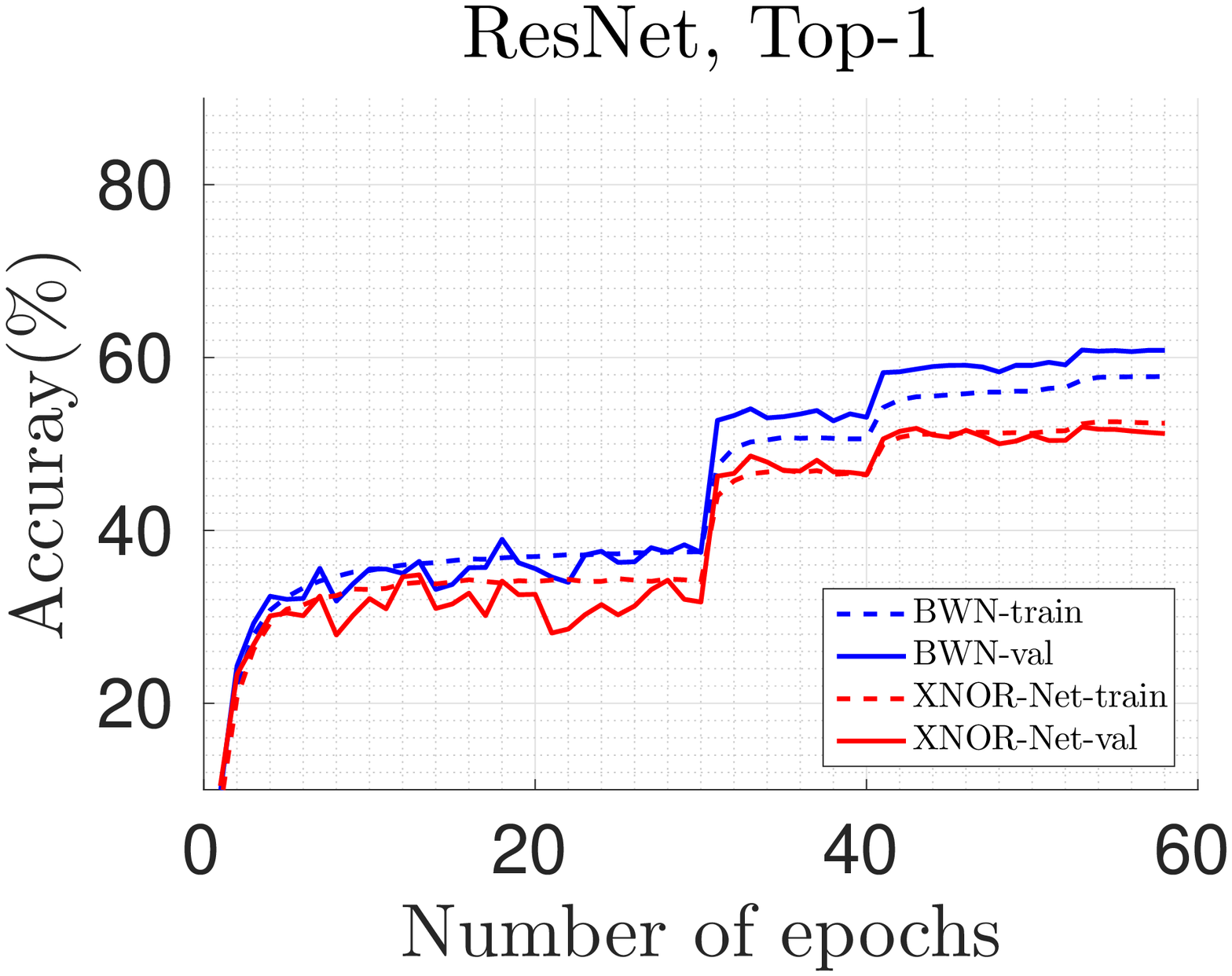}}
\subfloat[]{\includegraphics[height=0.25\textwidth]{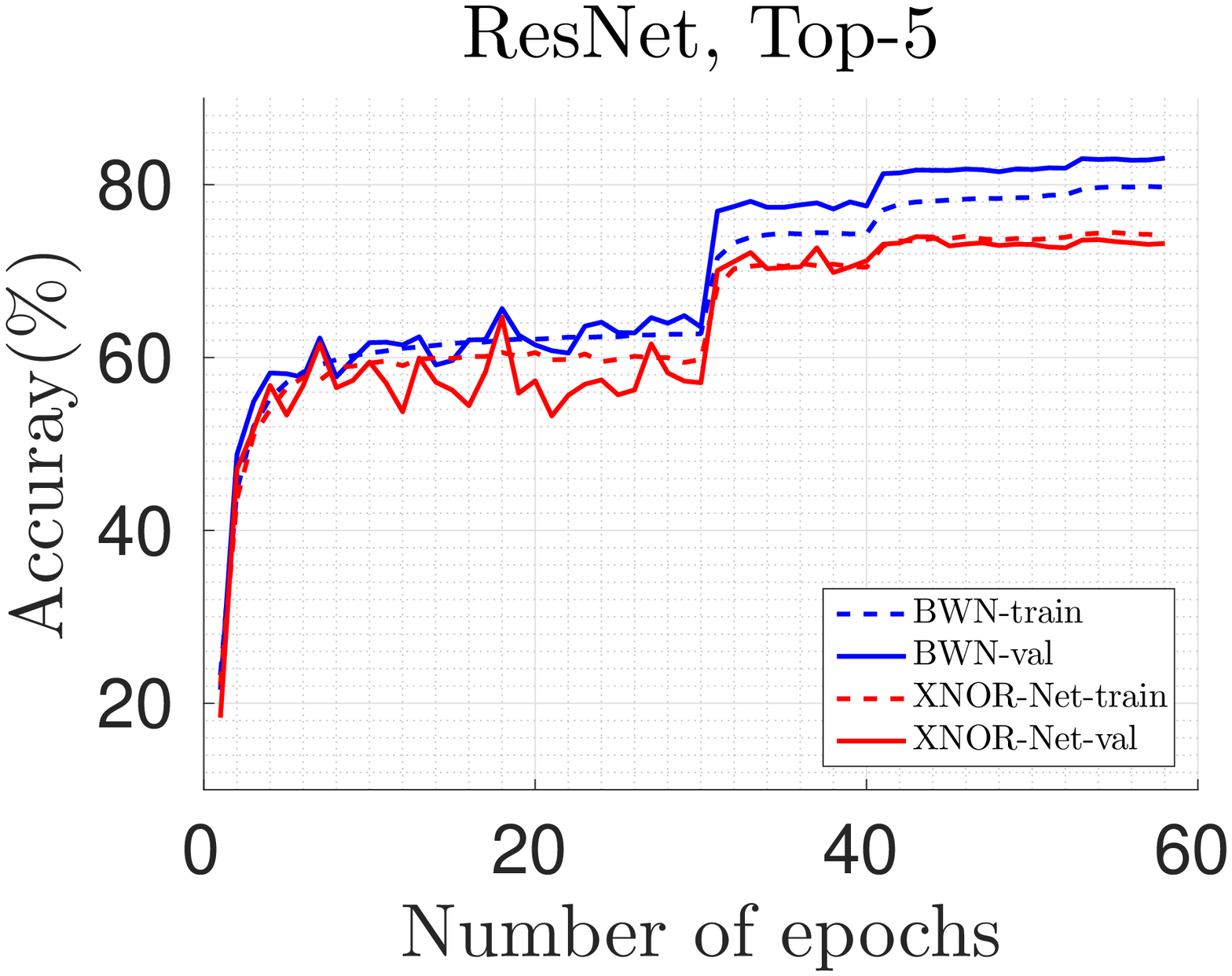}}
\caption{\footnotesize This figure shows the classification accuracy; (a)Top-1 and (b)Top-5 measures across the training epochs on ImageNet dataset by Binary-Weight-Network and XNOR-Network using ResNet-18.}
\label{fig:resnetcurve}
\end{figure*}

\begin{table}[t!]
\centering
\begin{tabular}{ |p{4cm}||p{1cm}|p{1cm} || p{1cm} | p{1cm}| }
 \hline
 & \multicolumn{2}{c||}{ResNet-18} & \multicolumn{2}{|c|}{GoogLenet} \\
 \hline
 Network Variations& top-1 & top-5 & top-1 & top-5\\
 \hline
 Binary-Weight-Network      &   60.8 &83.0 & 65.5 & 86.1 \\
 XNOR-Network&   51.2    &73.2 & N/A & N/A\\
 \hline
 \hline
 Full-Precision-Network&            69.3    &89.2 & 71.3 & 90.0\\ 
 \hline
\end{tabular}
\caption{\footnotesize This table compares the final classification accuracy achieved by our binary precision networks with the full precision network in ResNet-18 and GoogLenet architectures.}
\label{table:resnetcomp}
\end{table}
\textbf{Residual Net :}
We use the ResNet-18 proposed in ~\cite{kaming2016residual}  with short-cut type B.\footnote{We used the Torch implementation in https://github.com/facebook/fb.resnet.torch}         

\textit{Train:} In each training iteration, images are resized randomly between 256 and 480 pixel on the smaller dimension and then a random crop of $224\times 224$ is selected for training.  We run the training algorithm for 58 epochs with batch size equal to 256 images. The learning rate starts at 0.1 and we use the learning-rate-decay equal to 0.01 at epochs number 30 and 40.    

\textit{Test:} At inference time, we use the $224\times 224$ center crop for forward propagation. 

Figure~\ref{fig:resnetcurve} demonstrates the classification accuracy (Top-1 and Top-5) along the epochs for training and inference. The dashed lines represent training and the solid lines represent inference.  Table~\ref{table:resnetcomp} shows our final accuracy by BWN and XNOR-Net.    

\textbf{GoogLenet Variant :}
 We experiment with a variant of GoogLenet \cite{szegedy2015going} that uses a similar number of parameters and connections but only straightforward convolutions, no branching\footnote{We used the Darknet \cite{darknet13} implementation: http://pjreddie.com/darknet/imagenet/\#extraction}. It has 21 convolutional layers with filter sizes alternating between $1\times1$ and $3\times3$.

\textit{Train:} Images are resized randomly between 256 and 320 pixel on the smaller dimension and then a random crop of $224\times 224$ is selected for training.  We run the training algorithm for 80 epochs with batch size of 128. The learning rate starts at 0.1 and we use polynomial rate decay, $\beta = 4$.

\textit{Test:} At inference time, we use a center crop of $224\times 224$.

\subsection{Ablation Studies}
\begin{table}[t!]
\centering
\subfloat[][]{
\begin{tabular}{ |p{3.5cm}||p{1cm}|p{1cm}|  }
 \hline
 \multicolumn{3}{|c|}{Binary-Weight-Network} \\
 \hline
 Strategy for computing $\alpha$ & top-1 & top-5 \\
 \hline
 Using equation~\ref{eq:optalpha2}   & 56.8    &79.4\\
 Using a separate layer                  &46.2    &69.5\\
 \hline
\end{tabular}
}
\subfloat[][]{
\begin{tabular}{ |p{3cm}||p{1cm}|p{1cm}|  }
 \hline
 \multicolumn{3}{|c|}{XNOR-Network} \\
 \hline
 Block Structure& top-1 & top-5 \\
 \hline
 C-B-A-P   &30.3    &57.5\\
 B-A-C-P   &44.2    &69.2\\
 \hline
\end{tabular}
}
\caption{\footnotesize In this table, we evaluate two key elements of our approach; computing the optimal scaling factors and specifying the right order for layers in a block of CNN with binary input. (a) demonstrates the importance of the scaling factor in training binary-weight-networks and (b) shows that our way of ordering the layers in a block of CNN is crucial for training XNOR-Networks. C,B,A,P stands for Convolutional, BatchNormalization, Acive function (here binary activation), and Pooling respectively.}
\label{table:ablation}
\end{table}\
There are two key differences between our method and the previous network binariazation methods; the binararization technique and the block structure in our binary CNN. For binarization, we find the optimal scaling factors at each iteration of training. For the block structure, we order the layers in a block in a way that decreases the quantization loss for training XNOR-Net. Here, we evaluate the effect of each of these elements in the performance of the binary networks. 
Instead of computing the scaling factor $\alpha$ using equation~\ref{eq:optalpha2}, one can consider $\alpha$ as a network parameter. In other words, a layer after binary convolution multiplies the output of convolution by an scalar parameter for each filter. This is similar to computing the affine parameters in batch normalization.   Table~\ref{table:ablation}-a compares the performance of a binary network with two ways of computing the scaling factors. As we mentioned in section~\ref{sec:binconv} the typical block structure in CNN is not suitable for binarization. Table~\ref{table:ablation}-b compares the standard block structure C-B-A-P (Convolution, Batch Normalization, Activation, Pooling) with our structure B-A-C-P. (A, is binary activation).

\section{Conclusion}
\label{sec:conclusion}
We introduce simple, efficient, and accurate binary approximations for neural networks. We train a neural network that learns to find binary values for weights, which reduces the size of network by $\sim32\times$ and provide the possibility of loading very deep neural networks into portable devices with limited memory. We also propose an architecture, XNOR-Net, that uses mostly bitwise operations to approximate convolutions.  This provides $\sim58\times$ speed up and enables the possibility of running the inference of state of the art deep neural network on CPU (rather than GPU) in real-time.

\section*{Acknowledgements}
This work is in part supported by ONR N00014-13-1-0720, NSF IIS- 1338054, Allen
Distinguished Investigator Award, and the Allen Institute for Artificial Intelligence.

\bibliographystyle{splncs}
\bibliography{xnor}
\end{document}